\documentclass{article}
\usepackage{ijcai26}

\usepackage{times}
\usepackage{soul}
\usepackage{url}
\usepackage[hidelinks]{hyperref}
\usepackage[utf8]{inputenc}
\usepackage[small]{caption}
\usepackage{graphicx}
\usepackage{amsmath}
\usepackage{amssymb}
\usepackage{amsthm}
\usepackage{booktabs}
\usepackage{algorithm}
\usepackage{algorithmic}
\newcommand{\Comment}[1]{\hfill{\footnotesize$\triangleright$~#1}}
\usepackage[table]{xcolor}
\usepackage{colortbl}
\usepackage{subcaption}
\usepackage{booktabs}
\usepackage{multirow}
\usepackage[table]{xcolor}
\usepackage{makecell}

\title{Breaking the Pre-Sampling Barrier:\\Activation-Informed Difficulty-Aware Self-Consistency}

\author{
    Taewoong Yoon, Geunyeong Jeong, Geon Park, Sihyeong Yeom, Harksoo Kim\thanks{Corresponding author.}
    \affiliations
    Konkuk University
    \emails
    {\{twyoon816, jyjg7218, albert0811, stv10121, nlpdrkim}\}@konkuk.ac.kr
}

\begin{document}

\maketitle

\begin{abstract}
Self-Consistency (SC) is an effective decoding strategy that improves the reasoning performance of Large Language Models (LLMs) by generating multiple chain-of-thought reasoning paths and selecting the final answer via majority voting. However, it suffers from substantial inference costs because it requires a large number of samples. To mitigate this issue, Difficulty-Adaptive Self-Consistency (DSC) was proposed to reduce unnecessary token usage for easy problems by adjusting the number of samples according to problem difficulty. However, DSC requires additional model calls and pre-sampling to estimate difficulty, and this process is repeated when applying to each dataset, leading to significant computational overhead. In this work, we propose Activation-Informed Difficulty-Aware Self-Consistency (ACTSC) to address these limitations. ACTSC leverages internal difficulty signals reflected in the feed-forward network neuron activations to construct a lightweight difficulty estimation probe, without any additional token generation or model calls. The probe dynamically adjusts the number of samples for SC and can be applied to new datasets without requiring pre-sampling for difficulty estimation. To validate its effectiveness, we conduct experiments on five benchmarks. Experimental results show that ACTSC effectively reduces inference costs while maintaining accuracy relative to existing methods.
\end{abstract}

\section{Introduction}

\begin{figure}[t]
    \centering
    \includegraphics[width=0.8\columnwidth]{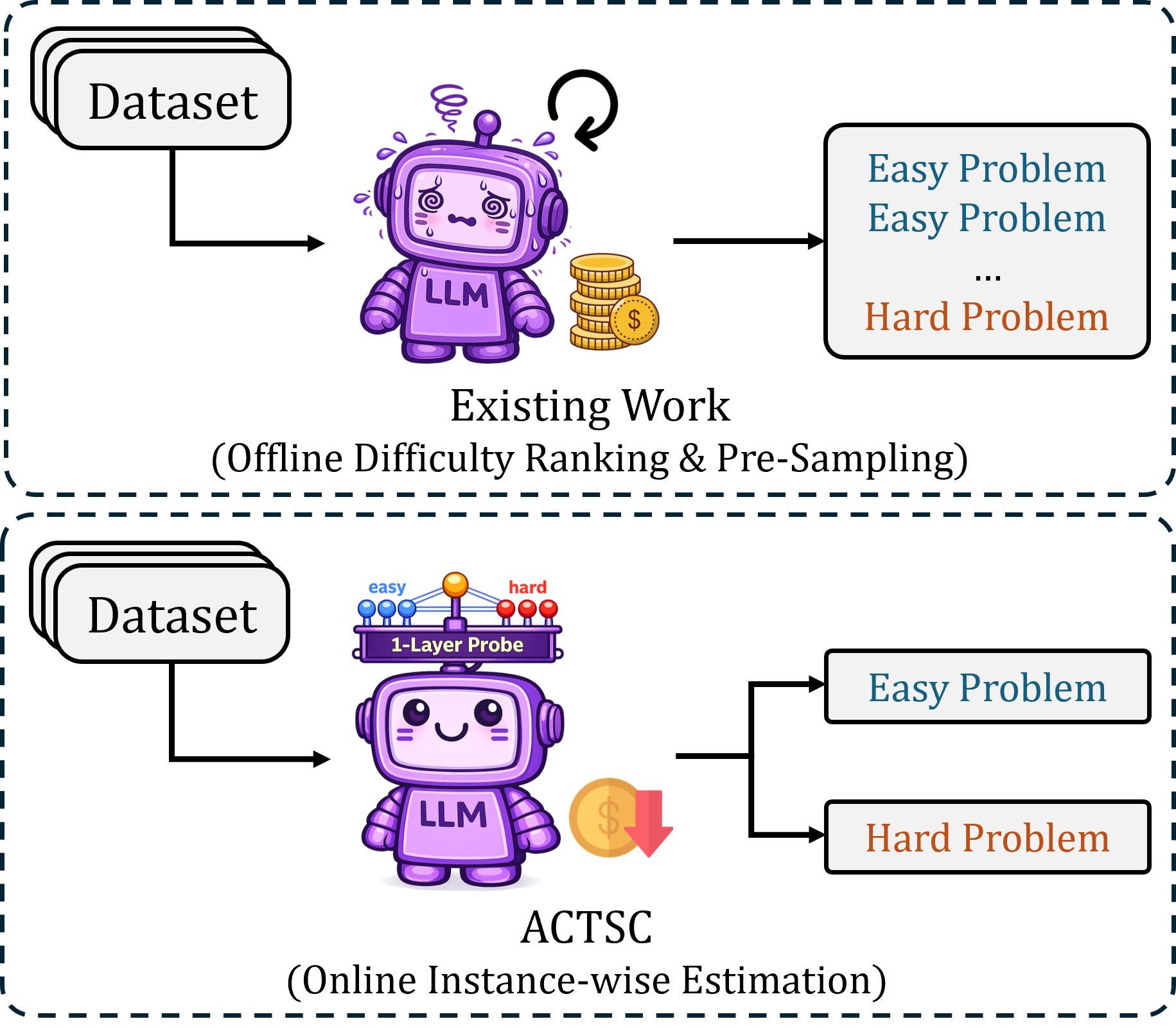}
    \caption{Comparison of existing methods and ACTSC. Existing methods need pre-sampling and extra model calls to estimate difficulty. ACTSC uses internal model signals from the input, eliminating additional token generation and model call costs.}
    \label{fig:neuron_gap}
\end{figure}

Large Language Models (LLMs) have improved their reasoning performance on complex tasks, such as mathematical, logical, and commonsense reasoning, through Chain-of-Thought (CoT) prompting~\cite{cot}. This performance gain is closely related to Test-Time Scaling (TTS)~\cite{tts} strategies, which improve performance by allocating more computational resources during inference. Among various TTS techniques, Self-Consistency (SC)~\cite{sc} has become the most widely used strategy. SC samples multiple reasoning paths for the same input and selects the final answer through majority voting, effectively reducing the instability of single CoT reasoning. However, SC has a fundamental limitation: it requires generating multiple samples, which leads to a sharp increase in inference costs.

To address this issue, recent research~\cite{ac,esc} has explored approaches that maintain SC's performance benefits while reducing inference costs. A representative method is Difficulty-Adaptive Self-Consistency (DSC)~\cite{dsc}. DSC ranks problems by difficulty level, then divides them into easy and hard categories based on where answer agreement becomes sufficiently high across consecutive problems. It then allocates computational resources adaptively by applying single reasoning to easy problems and multiple reasoning to hard ones. This aims to prevent unnecessary multiple reasoning on easy problems and improve overall inference efficiency. However, DSC requires a separate preparing stage for difficulty estimation, which must be repeated for each new dataset. When this cost is combined with inference stage costs, it can actually result in higher total costs than the original SC method on high-difficulty datasets. 

In this paper, we propose Activation-Informed Difficulty-Aware Self-Consistency (ACTSC), a framework designed to address these limitations. ACTSC first identifies difficulty-sensitive neurons within the model and uses them to train a lightweight difficulty estimation probe \cite{probe1,probe2}. Given an input problem, the trained probe estimates its difficulty using only a single forward pass, and the model subsequently adapts its inference budget according to the estimated difficulty. Unlike existing methods, our approach does not require a preliminary process for difficulty estimation on new datasets, and can perform difficulty-based computational resource allocation immediately during inference without additional pre-sampling or token generation. Experimental results across various benchmarks and models show that ACTSC reduces the number of samples by up to 87.1\% compared to SC and up to 40\% compared to DSC while maintaining performance. In this paper, our contributions are as follows:

\begin{itemize}
    \item We propose Activation-Informed Difficulty-Aware Self-Consistency (ACTSC), which adaptively allocates computational resources based on problem difficulty during inference by using internal activation signals from the model.
    \item We show that a lightweight difficulty estimation probe based on activation neurons can directly estimate difficulty at inference time without separate preliminary difficulty estimation or pre-sampling processes, even for new datasets.
    \item Through experiments on various reasoning benchmarks, we demonstrate that the proposed ACTSC significantly improves inference efficiency while maintaining stable accuracy compared to SC and its variants.
\end{itemize}

\section{Related Work}
\subsection{Self-Consistency-Based Test-Time Scaling}
Self-Consistency (SC)~\cite{sc} was proposed as an inference strategy that samples multiple Chain-of-Thought (CoT) reasoning paths for the same input and selects the final answer through agreement among them~\cite{cot}. The main limitation of SC is the high inference cost that comes from needing to generate multiple reasoning paths. To address this, previous studies have proposed various improved methods that control SC's sampling process more efficiently.

\begin{itemize}
    \item \textbf{Adaptive Consistency (AC)} gradually observes the level of agreement among responses generated during inference and stops additional sampling early when predefined stopping criteria are met \cite{ac}.
\item \textbf{Early-Stopping Self-Consistency (ESC)} performs inference in fixed-size sample windows and stops generating more samples when responses within a window show sufficient consistency \cite{esc}.
\item \textbf{Difficulty-Adaptive Self-Consistency (DSC)} proposes an approach that assigns different sampling budgets for each problem based on problem difficulty. This method aims to reduce average inference cost by using problem difficulty \cite{dsc}. However, it has the limitation that the difficulty estimation process requires multiple model calls to rank problem difficulty and incurs additional sampling costs to partition easy and hard problems.
\end{itemize}

\subsection{Difficulty Estimation}
Previous studies on estimating problem difficulty have treated difficulty as a latent variable that cannot be directly observed and have tried to infer it through various indirect signals. Recent studies on large language models have also extended this perspective and proposed several experimental approaches for difficulty estimation. Some studies have proposed methods that estimate problem difficulty by analyzing model generated outputs or intermediate ~\cite{lee2025semantic,esc}. While these methods are intuitive in that they use signals observed from the model's actual reasoning process, they often require generating multiple outputs or repeated reasoning steps for difficulty judgment, which leads to additional test-time computational costs.
Another research direction uses auxiliary large language models as judges or fine-tunes them to predict difficulty in order to evaluate problem difficulty more explicitly~\cite{dsc,cheng2025think}. While this approach evaluates difficulty by taking question text or limited reasoning results as input, it relies on judgments from a different model than the target model, which can create discrepancies with the difficulty perceived by the model actually performing the reasoning. Additionally, introducing auxiliary models increases system complexity and incurs additional inference costs.

\begin{figure*}[t]
    \centering
    \includegraphics[width=\textwidth]{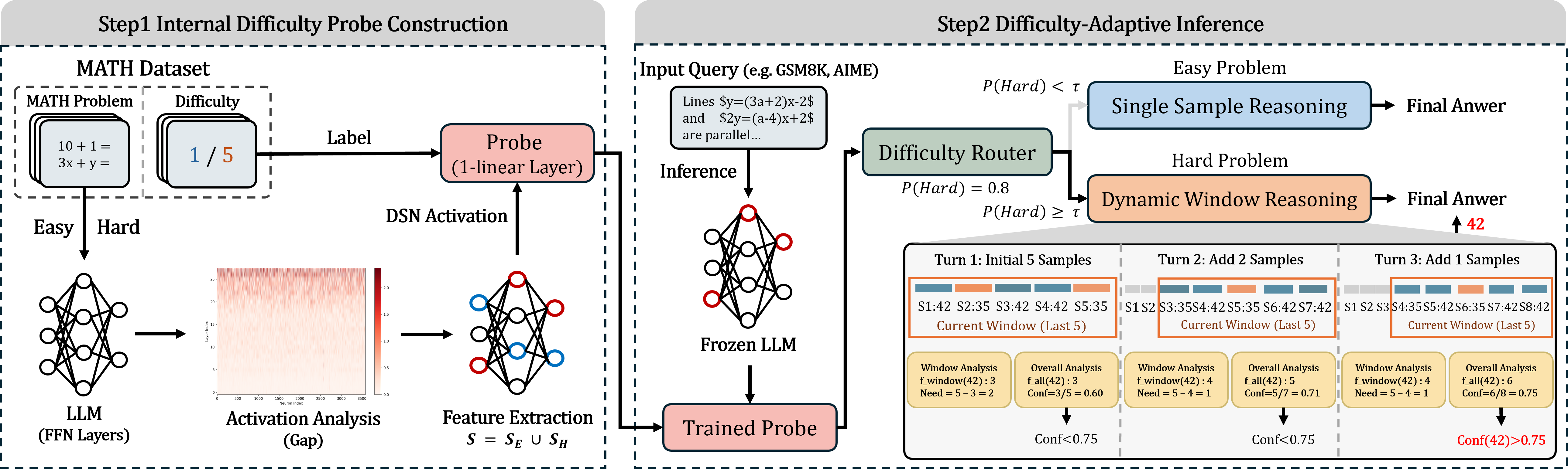}
    \caption{Overview of the ACTSC framework. ACTSC consists of two sequential steps. (Left) In Step 1, a lightweight difficulty probe is trained based on internal FFN activation patterns while keeping the LLM frozen. (Right) In Step 2, the trained probe estimates problem difficulty from activation signals observed during a forward pass and adaptively controls the number of inference samples by selecting either single-sample reasoning or dynamic window reasoning.}
    \label{fig:overall_framework}
\end{figure*}

\section{Methodology}

\begin{algorithm}[tb]
    \caption{Internal Difficulty Probe Construction}
    \label{alg:offline}
    \textbf{Input}: Dataset $\mathcal{D} = \{(q_i, d_i)\}_{i=1}^{N}$ with difficulty labels $d_i$, frozen LLM $M$ \\
    \textbf{Parameters}: Easy threshold $\theta_{\text{easy}}$, Hard threshold $\theta_{\text{hard}}$ \\
    \textbf{Output}: Trained difficulty probe $P: \mathbb{R}^{|\mathcal{S}|} \rightarrow [0,1]$
    \begin{algorithmic}[1]
        \STATE Run the frozen model to obtain FFN activations for each input:
        \STATE \hspace{0.5cm} $\mathbf{z}_i \leftarrow M(q_i)$
        
        \STATE Identify difficulty-sensitive neurons:
        \STATE \hspace{0.5cm} $\mathcal{S}_{\text{easy}} \leftarrow \{ n \mid \text{gap}(n, \theta_{\text{easy}}) > 0 \}$
        \STATE \hspace{0.5cm} $\mathcal{S}_{\text{hard}} \leftarrow \{ n \mid \text{gap}(n, \theta_{\text{hard}}) > 0 \}$
        \STATE \hspace{0.5cm} $\mathcal{S} \leftarrow \mathcal{S}_{\text{easy}} \cup \mathcal{S}_{\text{hard}}$
        
        \STATE Extract DSN activation features:
        \STATE \hspace{0.5cm} $\mathbf{h}_i \leftarrow \{ a_n(q_i) \mid n \in \mathcal{S} \}$
        
        \STATE Construct binary difficulty labels:
        \STATE \hspace{0.5cm} $y_i \leftarrow \mathbb{I}[d_i \ge \theta_{\text{hard}}]$
        
        \STATE Train a linear probe $P$ on $\{(\mathbf{h}_i, y_i)\}_{i=1}^{N}$ using binary cross-entropy loss
        
        \STATE \textbf{return} $P$
    \end{algorithmic}
\end{algorithm}

In this section, we describe the ACTSC methodology, which estimates the difficulty of an input problem using only internal FFN activations and dynamically controls the number of SC inference samples based on that difficulty. As shown in Figure~\ref{fig:overall_framework}, the overall ACTSC framework consists of two phases: difficulty probe construction and difficulty-adaptive inference, with each phase summarized in Algorithm~\ref{alg:offline} and Algorithm~\ref{alg:online}.

\subsection{Internal Difficulty Probe Construction}

The goal of the step1 is to train a lightweight difficulty probe that takes internal FFN activation signals from the language model as input and predicts the difficulty of the input problem. This phase is performed only once step1 using a MATH dataset~\cite{math} with difficulty labels
\begin{equation}
\mathcal{D} = \{(q_i, d_i)\}_{i=1}^{N}.
\end{equation}
Once trained, the probe is used during the inference phase without any additional training.

\subsubsection{Problem Setup}

We formulate difficulty estimation as a binary classification problem. Each problem is assigned a difficulty label $d_i \in \{1,\dots,5\}$, where the lowest difficulty (Level 1) is defined as easy and the highest difficulty (Level 5) as hard. We set two difficulty thresholds $\theta_{\text{easy}}$ and $\theta_{\text{hard}}$ and construct binary difficulty labels $y_i \in \{0,1\}$ as follows:
\begin{equation}
y_i = 
\begin{cases}
0, & \text{if } d_i \le \theta_{\text{easy}},\\
1, & \text{if } d_i \ge \theta_{\text{hard}}.
\end{cases}
\end{equation}
Samples in the intermediate difficulty range are excluded from probe training to reduce label uncertainty and representation entanglement.

\subsubsection{Identifying Difficulty-Sensitive Neurons (DSN)}

For each input problem, we run the language model through a single forward pass to extract internal FFN activations. In this work, we use the FFN activation corresponding to the last token of the input sequence as the analysis target.

To identify difficulty-sensitive neurons, we define a function $\text{gap}(n,\theta)$ that measures the degree of activation separation based on a difficulty threshold for each FFN neuron. This function represents the difference in average activation between two difficulty groups divided by the threshold and serves as an indicator to identify neurons that are selectively activated for specific difficulty categories.
\begin{equation}
\text{gap}(n,\theta) = \mathbb{E}[a_n(q)\mid d \le \theta] - \mathbb{E}[a_n(q)\mid d > \theta]
\end{equation}
Based on this, we define two neuron sets:
\begin{equation}
\begin{aligned}
\mathcal{S}_{\text{easy}} &= \{n \mid \text{gap}(n,\theta_{\text{easy}}) > 0\},\\
\mathcal{S}_{\text{hard}} &= \{n \mid \text{gap}(n,\theta_{\text{hard}}) > 0\}.
\end{aligned}
\end{equation}
Finally, we define the union of these two sets
\begin{equation}
\mathcal{S} = \mathcal{S}_{\text{easy}} \cup \mathcal{S}_{\text{hard}}.
\end{equation}
as Difficulty-Sensitive Neurons (DSN). This set is highly sparse relative to the full FFN neuron space and contains only the core neurons that directly contribute to difficulty discrimination.

\subsubsection{Training a Linear Difficulty Probe}

Having identified the DSN set $\mathcal{S}$, we now use these neurons as input features to train a lightweight estimator for difficulty prediction. Specifically, for each input problem, we input it into the language model and extract a feature vector by collecting only the activation values of neurons $n \in \mathcal{S}$:
\begin{equation}
\mathbf{h}_i = \{a_n(q_i) \mid n \in \mathcal{S}\} \in \mathbb{R}^{|\mathcal{S}|}.
\end{equation}
To normalize for differences in activation magnitude across neurons, we apply normalization to the feature vector prior to probe training.

We train a difficulty probe in the form of a linear estimator for difficulty prediction. The probe consists of a weight vector $\mathbf{w}$ and bias $b$, and outputs the probability that the input problem belongs to the hard difficulty category through a sigmoid function:
\begin{equation}
\hat{y}_i = P(\text{Hard}\mid q_i) = \sigma(\mathbf{w}^{\top}\mathbf{h}_i + b).
\end{equation}

The probe is trained using the binary difficulty labels $y_i$, with optimization performed by minimizing the Binary Cross-Entropy (BCE) loss between the actual labels and predicted probabilities:
\begin{equation}
\mathcal{L} = -\frac{1}{N}\sum_{i=1}^{N}\left[y_i\log(\hat{y}_i) + (1-y_i)\log(1-\hat{y}_i)\right].
\end{equation}
During training, all language model parameters remain frozen, with only the probe parameters being updated. This design ensures that difficulty prediction focuses on decoding signals already encoded in DSN activations, rather than learning new representations. Once training is complete, the difficulty probe can estimate problem difficulty from a single forward pass during inference.

\subsection{Difficulty-Adaptive Inference}

\begin{algorithm}[tb]
    \caption{Difficulty-Adaptive Inference}
    \label{alg:online}
    \textbf{Input}: Query $q$, LLM $M$, Trained probe $P$\\
    \textbf{Parameter}: Difficulty threshold $\tau$, Confidence threshold $\gamma$, Window size $w$\\
    \textbf{Output}: Final answer $\hat{a}$
    \begin{algorithmic}[1]
        \STATE Extract DSN activations $a \leftarrow \text{ExtractDSN}(M(q), S)$
        \STATE Predict difficulty: $P_{hard} \leftarrow P(a)$
        \IF{$P_{hard} < \tau$}
            \STATE \textbf{return} $M(q)$ \Comment{Single Sample Reasoning}
        \ELSE
            \STATE $\mathcal{S} \leftarrow []$
            \REPEAT
                \STATE $n_{need} \leftarrow w - \max_{a'} f_{window}(a', \mathcal{S}[-w:])$
                \STATE Generate $n_{need}$ samples: $\mathcal{S} \leftarrow \mathcal{S} \cup \{M(q)\}_{i=1}^{n_{need}}$
                \STATE $a^* \leftarrow \arg\max_{a'} \text{Count}(a', \mathcal{S}) / |\mathcal{S}|$
            \UNTIL{$\text{Conf}(a^*) \geq \gamma$}
            \STATE \textbf{return} $a^*$ \Comment{Dynamic window reasoning}
        \ENDIF
    \end{algorithmic}
\end{algorithm}

In the online phase, the trained difficulty probe is used to immediately estimate the difficulty of each input problem and dynamically select an inference strategy accordingly. The key idea of this phase is to secure both computational efficiency and inference stability by performing additional sampling only when necessary based on the problem difficulty distribution, instead of applying the same SC inference to all inputs.

\subsubsection{Difficulty Prediction}

Given an input query $q$, we run the language model once to extract FFN activation signals corresponding to the DSN set. This activation vector is used as input to the difficulty probe trained in the step 1, which computes the probability that the input problem belongs to the hard difficulty category: $P_{\text{hard}} = P(\text{Hard}\mid q)$. Since this process is based on a single forward pass, it requires no additional sampling or repeated inference. Therefore, the computational cost incurred in the difficulty estimation step is nearly identical to single-sample inference and serves as a lightweight decision signal for subsequent inference strategy selection.

\subsubsection{Difficulty-Based Routing}

The estimated difficulty probability $P_{\text{hard}}$ is compared with a difficulty threshold $\tau$ to determine the inference path. In this work, we set $\tau$ as the average value of $P_{\text{hard}}$ observed in each dataset. That is, we define $\tau = \mathbb{E}_{q \sim \mathcal{D}}[P(\text{Hard}\mid q)]$, which represents the baseline corresponding to an ``averagely difficult problem'' in that dataset. This setup allows ACTSC to naturally reflect differences in difficulty distributions across datasets, instead of using a fixed threshold. As a result, ACTSC selectively applies SC inference only to relatively difficult problems in a specific dataset. 

Specifically, when $P_{\text{hard}} < \tau$, the input problem is judged to be relatively easy, and the single inference result from the language model is used directly as the final answer. Conversely, when $P_{\text{hard}} \ge \tau$, we judge that single inference alone is likely to produce unstable answers and proceed with the Self-Consistency-based multiple sampling procedure. This difficulty-based routing prevents unnecessary sampling for easy problems while allocating additional inference resources only to difficult problems.

\subsubsection{Dynamic Window Reasoning}

For problems judged as hard, we apply a dynamic window reasoning strategy instead of the conventional SC approach that uses a fixed number of samples. Let $\mathcal{S}$ denote the set of answers generated during the inference process. We maintain a sliding window consisting of the most recent $w$ answers and continuously track the most frequent answer within that window. At each iteration, we calculate the number of additional samples needed $n_{\text{need}}$ based on the frequency of the current most frequent answer in the window and generate only that many new inference samples. This allows inference to terminate without unnecessary additional sampling when the answer distribution has sufficiently converged. 

When the confidence $\text{Conf}(a^{*})$ of the most frequent answer $a^{*}$ in the window exceeds a predefined threshold $\gamma$, that answer is returned as the final output. This termination condition allows SC inference to stop based on convergence state rather than a fixed number of samples, enabling adaptive adjustment of the inference amount needed for each problem. This dynamic window reasoning improves computational efficiency compared to conventional fixed-sample SC while preventing performance degradation by continuing inference until answer stability is secured.

\begin{table*}[t]
\centering
\footnotesize
\renewcommand{\arraystretch}{1.0}
\setlength{\tabcolsep}{3pt}

\resizebox{\textwidth}{!}{%
\begin{tabular}{c|ccc|ccc|ccc}
\toprule
\multicolumn{1}{c|}{\multirow[c]{2}{*}{\textbf{Method}}}
& \multicolumn{3}{c|}{\textbf{MATH-500}}
& \multicolumn{3}{c|}{\textbf{AIME 2024}}
& \multicolumn{3}{c}{\textbf{AIME 2025}} \\
\cmidrule(lr){2-4}
\cmidrule(lr){5-7}
\cmidrule(lr){8-10}
& Sample $\downarrow$ & Prepare / Inference$\downarrow$ & Acc $\uparrow$
& Sample $\downarrow$ & Prepare / Inference$\downarrow$ & Acc $\uparrow$
& Sample $\downarrow$ & Prepare / Inference$\downarrow$ & Acc $\uparrow$ \\
\midrule

\rowcolor{gray!15}
\multicolumn{10}{c}{\textbf{Qwen2.5-3B}} \\
\midrule
SC  & 40.00 & -- / 22.3 & 61.80 
    & 40.00 & -- / 42.3 & 11.20 
    & 40.00 & -- / 35.6 & 6.67 \\
AC  & 15.00 (\textcolor{green!60!black}{-62.5\%}) & -- / 14.0 & 61.79 
    & 34.68 (\textcolor{green!60!black}{-13.3\%}) & -- / 44.1 & 11.20 
    & 36.73 (\textcolor{green!60!black}{-8.2\%}) & -- / 42.4 & 6.67 \\
ESC & 20.21 (\textcolor{green!60!black}{-49.5\%}) & -- / 15.0 & 61.81 
    & 38.81 (\textcolor{green!60!black}{-3.0\%}) & -- / 43.0 & 11.20 
    & 39.83 (\textcolor{green!60!black}{-0.4\%}) & -- / 37.8 & 6.67 \\
DSC & 17.44 (\textcolor{green!60!black}{-56.4\%}) & 2.8 / 12.4 & 61.79 
    & 37.85 (\textcolor{green!60!black}{-5.4\%}) & 4.9 / 41.3 & 11.20 
    & 37.90 (\textcolor{green!60!black}{-5.2\%}) & 4.9 / 35.1 & 6.67 \\
\rowcolor{blue!15}
\textbf{ACTSC}
    & \textbf{9.71} (\textcolor{green!60!black}{-75.7\%}) & -- / \textbf{8.6} & 61.60 
    & \textbf{28.17} (\textcolor{green!60!black}{-29.6\%}) & -- / \textbf{33.7} & \textbf{11.20}
    & \textbf{30.13} (\textcolor{green!60!black}{-24.7\%}) & -- / \textbf{29.4} & \textbf{10.00} \\

\midrule
\rowcolor{gray!15}
\multicolumn{10}{c}{\textbf{Qwen2.5-7B}} \\
\midrule
SC  & 40.00 & -- / 21.2 & 66.71
    & 40.00 & -- / 42.2 & 16.67 
    & 40.00 & -- / 37.3 & 16.67 \\
AC  & 11.32 (\textcolor{green!60!black}{-71.7\%}) & -- / 10.3 & 66.76
    & 32.30 (\textcolor{green!60!black}{-19.3\%}) & -- / 41.2 & 16.67
    & 32.58 (\textcolor{green!60!black}{-18.6\%}) & -- / 39.3 & 16.67 \\
ESC & 15.99 (\textcolor{green!60!black}{-60.0\%}) & -- / 11.7 & 66.76
    & 37.11 (\textcolor{green!60!black}{-7.2\%}) & -- / 41.5 & 16.67
    & 37.64 (\textcolor{green!60!black}{-5.9\%}) & -- / 37.4 & 16.67 \\
DSC & 17.50 (\textcolor{green!60!black}{-56.2\%}) & 0.4 / 9.2 & 66.72
    & 34.11 (\textcolor{green!60!black}{-14.7\%}) & 5.0 / 37.8 & 16.67
    & 33.86 (\textcolor{green!60!black}{-15.3\%}) & 5.2 / 32.7 & 16.67 \\
\rowcolor{blue!15}
\textbf{ACTSC}
    & \textbf{6.72} (\textcolor{green!60!black}{-83.2\%}) & -- / \textbf{5.7} & \textbf{69.60}
    & \textbf{24.70} (\textcolor{green!60!black}{-38.3\%}) & -- / \textbf{29.0} & \textbf{16.67}
    & \textbf{26.20} (\textcolor{green!60!black}{-34.5\%}) & -- / \textbf{27.6} & 13.33 \\

\midrule
\rowcolor{gray!15}
\multicolumn{10}{c}{\textbf{Gemma3-4B}} \\
\midrule
SC  & 40.00 & -- / 164.0 & 67.64 
    & 40.00 & -- / 163.8 & 13.33 
    & 40.00 & -- / 164.1 & 16.67 \\
AC  & 10.64 (\textcolor{green!60!black}{-73.4\%}) & -- / 46.0 & 67.64
    & 27.30 (\textcolor{green!60!black}{-31.8\%}) & -- / 116.7 & 13.33
    & 27.12 (\textcolor{green!60!black}{-32.2\%}) & -- / 119.1 & 16.67 \\
ESC & 14.40 (\textcolor{green!60!black}{-64.0\%}) & -- / 59.7 & 67.64
    & 33.70 (\textcolor{green!60!black}{-15.7\%}) & -- / 139.3 & 13.33
    & 32.72 (\textcolor{green!60!black}{-18.2\%}) & -- / 136.3 & 16.67 \\
DSC & 11.78 (\textcolor{green!60!black}{-70.5\%}) & 4.9 / 48.5 & 67.64 
    & 29.78 (\textcolor{green!60!black}{-25.5\%}) & 17.0 / 122.5 & 13.33
    & 28.96 (\textcolor{green!60!black}{-27.6\%}) & 17.7 / 119.8 & 16.67 \\
\rowcolor{blue!15}
\textbf{ACTSC}
    & \textbf{5.17} (\textcolor{green!60!black}{-87.1\%}) & -- / \textbf{21.5} & \textbf{70.00}
    & \textbf{18.80} (\textcolor{green!60!black}{-53.0\%}) & -- / \textbf{78.1} & 10.00
    & \textbf{17.90} (\textcolor{green!60!black}{-55.2\%}) & -- / \textbf{74.5} & \textbf{16.67} \\

\bottomrule
\end{tabular}
}
\caption{Performance comparison of various reasoning methods on mathematical reasoning benchmarks. We compare Self-Consistency (SC), Adaptive Consistency (AC), Early-Stopping Self-Consistency (ESC), and Difficulty-Adaptive Self-Consistency (DSC). Relative sample reduction vs.\ SC is reported in parentheses next to the Sample count.}
\label{tab:main_results_math}
\end{table*}

\section{Experiments}

\subsection{Experimental Setup}

\subsubsection{Benchmarks}
To evaluate the generalization capabilities of ACTSC, we employ both mathematical and non-mathematical reasoning benchmarks, covering diverse domains and difficulty levels. For mathematical reasoning, we use MATH-500~\cite{math}, a curated subset of 500 advanced problems from the MATH dataset, and AIME 2024~\cite{AIME2024} and AIME 2025~\cite{AIME2025}, each containing 30 high-difficulty problems from the American Invitational Mathematics Examination, which are more challenging than MATH. For non-mathematical reasoning, we include GPQA-Diamond~\cite{gpqa}, a benchmark for expert-level scientific knowledge,and MMLU-Pro~\cite{mmlu}, which covers diverse domains of expert-level world knowledge. In this experiment, we randomly sample 100 questions from MMLU-Pro due to its scale, using a fixed random seed of 42, resulting in 1,400 evaluation instances in total.

\subsubsection{Models}
In this experiment, we use three instruction-tuned LLMs: Gemma3-it (4B)~\cite{gemma} and Qwen2.5-Instruct (3B \& 7B)~\cite{qwen2.5}. These models differ in architecture and scale, allowing us to examine the robustness of the experimental results across diverse model configurations.

\subsubsection{Methods}
We compare ACTSC against four SC-based methods. SC uses a fixed sampling budget of $k = 40$, while the remaining methods dynamically determine the number of samples based on their stopping criteria, with a maximum budget of $k = 40$. All methods use the same decoding configuration, with a temperature of 0.7 and top-$p$ set to 0.8.

\begin{itemize}
\item \textbf{SC}~\cite{sc} generates a fixed set of $k = 40$ reasoning paths and selects the final answer via majority voting.

\item \textbf{AC}~\cite{ac} generates reasoning paths sequentially and terminates sampling when the agreement ratio exceeds 0.95.

\item \textbf{ESC}~\cite{esc} samples reasoning paths in windows of size $w = 5$ and stops when all answers within a window are identical.

\item \textbf{DSC}~\cite{dsc} estimates problem difficulty via pre-sampling; generates a single sample for easy problems and adaptively samples for hard problems until the agreement ratio reaches 0.95.

\item \textbf{ACTSC (Ours)} predicts problem difficulty using a trained probe without pre-sampling; for easy problems ($P_{\text{hard}} < \tau$), a single sample is generated, while for hard problems ($P_{\text{hard}} \ge \tau$), sampling is performed in windows of size $w = 5$ and terminates when the confidence of the most frequent answer exceeds $\gamma = 0.50$.
\end{itemize}

\subsubsection{Metrics}
In this study, we use the following evaluation metrics. Sample denotes the average number of response samples generated per problem during inference. Token cost is measured as the average number of input and output tokens and is divided into the preparation and inference phases. Prepare represents the token cost incurred during the pre-processing stage required by the DSC method for difficulty estimation, while Inference refers to the token cost used during the inference stage. All token costs are reported in thousands (k). The sample count and token cost metrics are used to evaluate inference efficiency. Accuracy (Acc) denotes the proportion of correctly answered problems on each benchmark and is used as a metric to evaluate the accuracy of each method.

\subsection{Experimental Results}

\subsubsection{Main Results}
Table~\ref{tab:main_results_math} summarizes the efficiency and accuracy results on mathematical reasoning benchmarks. The results show that ACTSC substantially improves efficiency while maintaining accuracy comparable to existing self-consistency variants. In particular, ACTSC achieves higher efficiency than DSC, which also incorporates difficulty information, by requiring fewer samples while maintaining comparable accuracy and eliminating preparation overhead. For instance, on AIME 2025 with Gemma3-4B, ACTSC reduces the average number of samples by 38.2\% (28.96 $\rightarrow$ 17.90) and the total token cost by 45.8\% (137.5k $\rightarrow$ 74.5k) relative to DSC, while maintaining comparable accuracy.

These efficiency gains are consistently observed across models of different scales and mathematical benchmarks spanning a wide range of difficulty levels, including high-difficulty AIME datasets. This consistency indicates that the proposed ACTSC framework, which leverages activation-informed difficulty estimation, provides appropriate sampling budgets across diverse models and problem distributions.

\begin{figure}[t]
    \centering
    \includegraphics[width=\columnwidth]{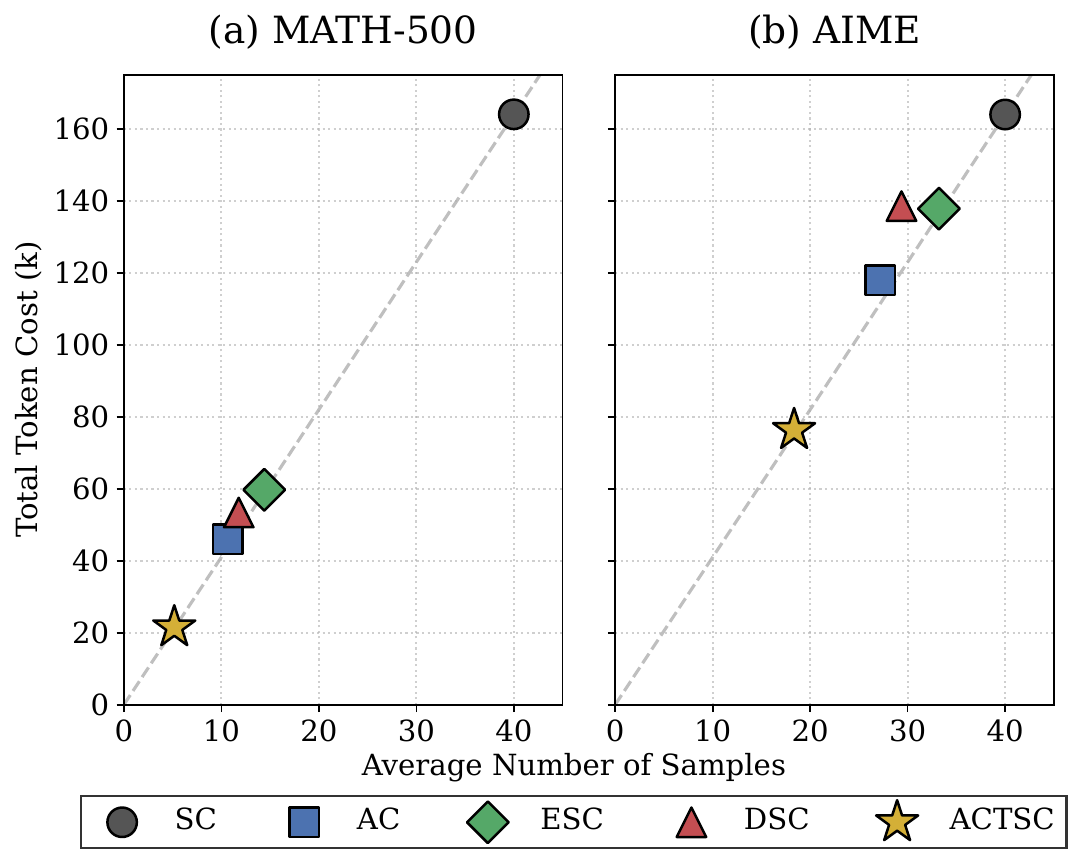}
    \caption{Total token cost vs. average number of samples for Gemma3-4B on (a) MATH-500 and (b) AIME average. The dashed line indicates a linear scaling baseline. Markers represent adaptive consistency methods.}
    \label{fig:efficiency_scaling}
\end{figure}

\subsubsection{Performance on Hard Benchmarks}
Reducing inference cost becomes increasingly difficult as benchmark difficulty rises, since a larger fraction of hard problems typically requires more sampling to stabilize reasoning. To evaluate efficiency under such conditions, we compare performance on a relatively easier benchmark (MATH-500) and a harder benchmark (AIME, averaged over 2024 and 2025). As shown in Figure~\ref{fig:efficiency_scaling}(a), several adaptive methods achieve noticeable efficiency gains over SC on MATH-500. However, Figure~\ref{fig:efficiency_scaling}(b) shows that these gains largely diminish on AIME, where most methods converge toward the high-cost behavior of SC.

In contrast, the proposed method maintains a clear efficiency advantage even on the harder benchmark, with the separation from other methods becoming more pronounced as difficulty increases. Notably, Table~\ref{tab:main_results_math} shows that DSC incurs substantially higher preparation cost as benchmark difficulty rises: its average preparation cost increases from 4.9k tokens on MATH-500 to 17.4k tokens on AIME, reflecting the growing expense of pre-difficulty estimation on harder datasets. The proposed method is unaffected by this trend, as it does not rely on any preparation stage, enabling efficient allocation of sampling budgets regardless of dataset difficulty.

\begin{table}[t]
\centering
\resizebox{0.9\columnwidth}{!}{

\begin{tabular}{c|ccc}
\toprule
\textbf{Method}
& Sample$\downarrow$
& Prepare / Inference$\downarrow$
& Acc$\uparrow$ \\
\midrule

\rowcolor{gray!15}
\multicolumn{4}{c}{\textbf{GPQA-Diamond}} \\
\midrule
SC   & 40.0 & -- / 19.6 & 36.67 \\
AC   & 18.4 (\textcolor{green!60!black}{-54.0\%}) & -- / 14.8 & 36.72 \\
ESC  & 24.7 (\textcolor{green!60!black}{-38.3\%}) & -- / 14.3 & 36.67 \\
DSC  & 23.1 (\textcolor{green!60!black}{-42.3\%}) & 3.4 / 12.2 & 36.66 \\
\rowcolor{blue!15}
\textbf{ACTSC}
     & \textbf{8.7} (\textcolor{green!60!black}{-78.3\%}) & -- / \textbf{6.2} & \textbf{38.46} \\

\midrule

\rowcolor{gray!15}
\multicolumn{4}{c}{\textbf{MMLU-Pro}} \\
\midrule
SC   & 40.0 & -- / 5.7 & 56.62 \\
AC   & 9.8  (\textcolor{green!60!black}{-75.5\%}) & -- / 5.3 & 56.59 \\
ESC  & 13.7 (\textcolor{green!60!black}{-65.8\%}) & -- / 4.2 & 56.59 \\
DSC  & 11.1 (\textcolor{green!60!black}{-72.3\%}) & 1.2 / 3.0 & 56.60 \\
\rowcolor{blue!15}
\textbf{ACTSC}
     & \textbf{6.6} (\textcolor{green!60!black}{-83.5\%}) & -- / \textbf{2.9} & 56.36 \\

\bottomrule
\end{tabular}
}
\caption{Performance on non-mathematical benchmarks (GPQA-Diamond and MMLU-Pro) using Qwen2.5-7B.}
\label{tab:main_results_nonmath}
\end{table}

\subsubsection{Generalization to Non-Mathematical Reasoning}
To assess generalization beyond mathematical reasoning, we evaluate the proposed method on two non-mathematical benchmarks using Qwen2.5-7B. As shown in Table~\ref{tab:main_results_nonmath}, the results align with those from mathematical benchmarks, with the proposed method consistently improving efficiency compared to other comparison methods while maintaining comparable accuracy. Specifically, compared to DSC, the proposed method reduces the average number of samples by 62.3\% (23.1 $\rightarrow$ 8.7) and the total token cost by 60.3\% (15.6k $\rightarrow$ 6.2k) on GPQA-Diamond, and by 40.5\% (11.1 $\rightarrow$ 6.6) and 31.0\% (4.2k $\rightarrow$ 2.9k) on MMLU-Pro, respectively. These consistent gains across domains indicate that the proposed activation-informed difficulty estimation generalizes beyond mathematical reasoning and enables domain-independent allocation of appropriate sampling budgets.

\subsection{Analysis}

\subsubsection{Probe Validity and Calibration}

\begin{figure}[t]
\centering
\includegraphics[width=\columnwidth]{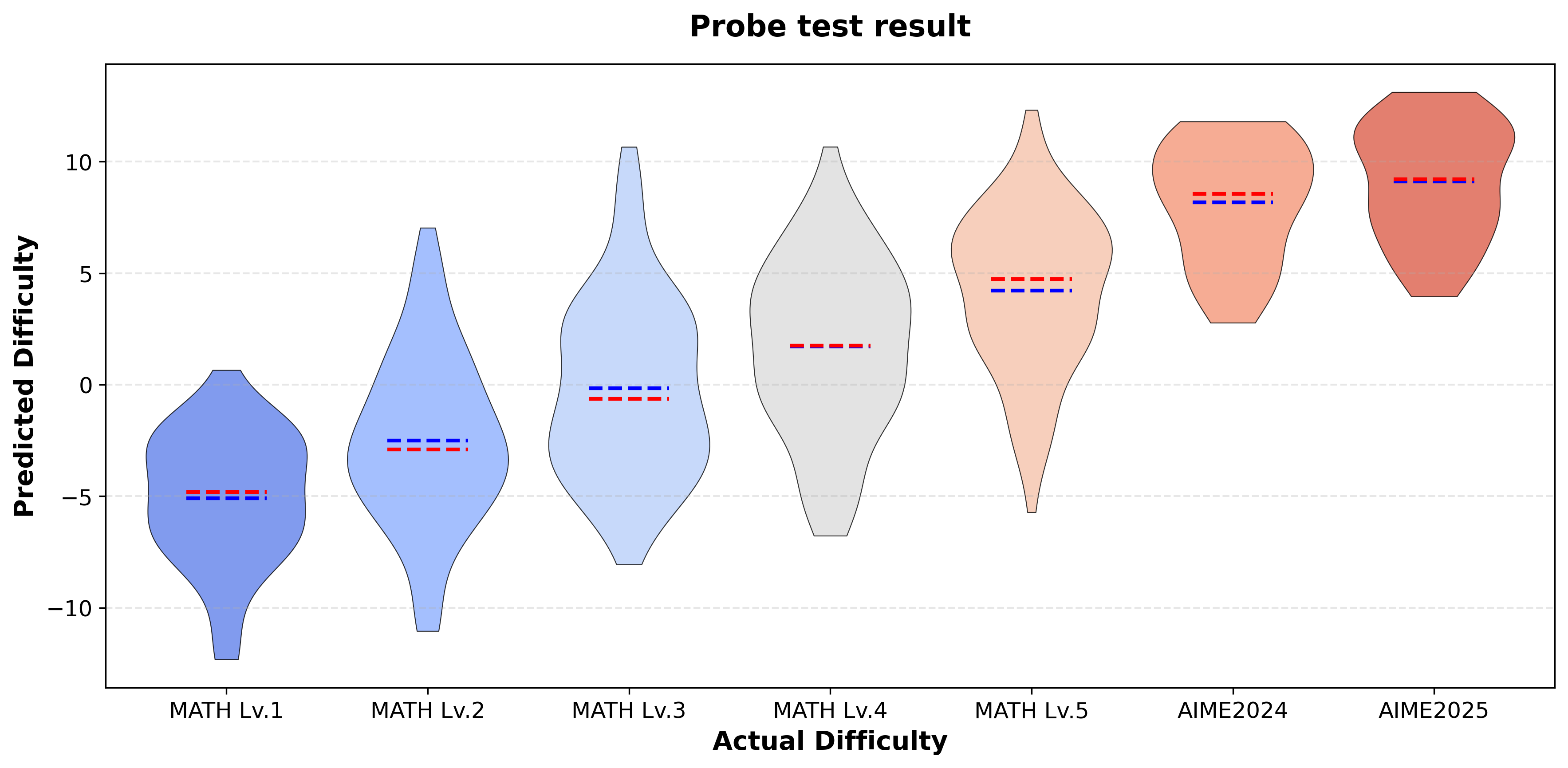}
\caption{Distribution of difficulty prediction logits produced by the DSN probe across different actual difficulty levels. Violin plots show the distribution of probe logits for MATH-500 Level 1–5 and AIME 2024/2025. Blue and red dashed lines represent the mean and median, respectively.}
\label{fig:probe_logits}
\end{figure}

We validate our proposed method through detailed analysis. To this end, we systematically examine the relationship between DSN-based probes and actual problem difficulty, and confirm that the signals extracted from internal representations are consistently aligned with true difficulty. Figure~\ref{fig:probe_logits} shows the distribution of logit values output by the trained probe across different actual difficulty ranges. A progressive increase in the distribution center of probe logits is observed from MATH Level 1 to Level 5, with AIME 2024 and 2025 problems concentrated in even higher logit regions than MATH Level 5. This indicates that the probe's logit outputs have been trained to reflect the relative magnitude of actual problem difficulty, demonstrating that difficulty-related signals can be linearly extracted from internal representations without token generation.

\begin{figure}[t]
\centering
\includegraphics[width=\columnwidth]{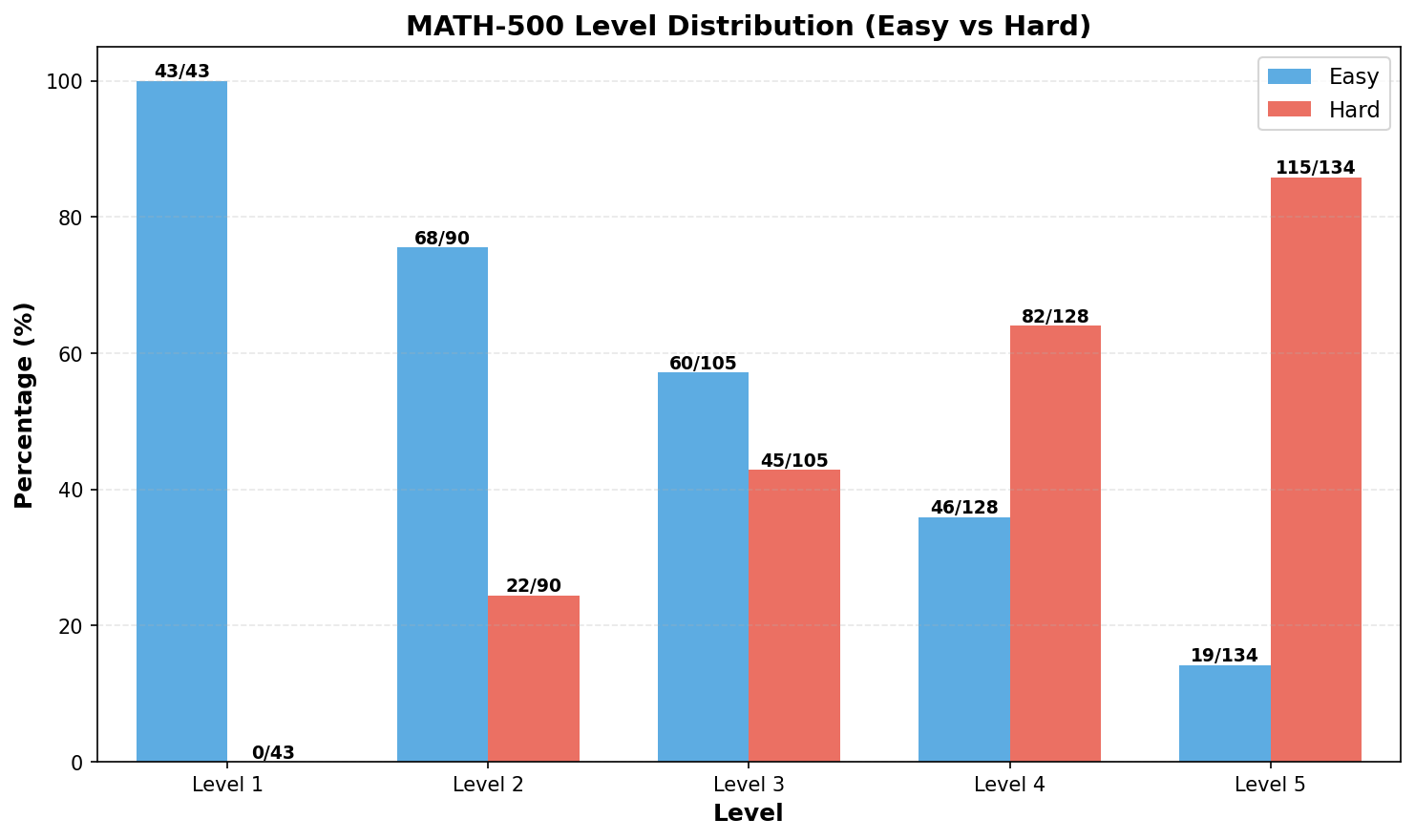}
\caption{Distribution of easy and hard problems across MATH difficulty levels using model-specific probes trained on internal activations of Qwen2.5-7B model. Percentages indicate the proportion of problems classified as easy or hard at each difficulty level.}
\label{fig:difficulty_distribution}
\end{figure}

Figure~\ref{fig:difficulty_distribution} presents the difficulty distribution results using a probe trained on internal activations from the Qwen2.5-7B model. As the MATH difficulty level increases, the model exhibits an ideal trend: the proportion of problems classified as easy decreases while the proportion classified as hard increases. This demonstrates that the probe trained on the model's activations accurately distinguishes difficulty levels across problems.

\subsubsection{Computational Overhead of the Difficulty Probe}

We analyze the computational overhead of the difficulty probe by distinguishing between a one-time offline cost and an inference-time cost. All measurements are conducted using the Qwen2.5-7B model with AIME 2025 as the reference dataset. The offline phase consists of difficulty-sensitive neuron (DSN) identification and linear probe training. DSN identification requires 138.7 seconds, and probe training adds 5.9 seconds, resulting in a total offline cost of 144.6 seconds. This cost is incurred once per model and is amortized across all subsequent inference runs.

At inference time, ACTSC estimates difficulty using FFN activations from a single forward pass without additional token generation. The measured overhead for computing $P(\text{Hard})$ is 0.59 seconds per sample, which is comparable to standard single-sample inference and negligible relative to multi-sample self-consistency. Unlike pre-sampling-based methods such as DSC, this overhead is independent of dataset size and task difficulty. Once trained, the probe introduces a fixed and predictable cost while enabling substantial reductions in self-consistency sampling.

\section{Conclusion}

We analyzed the inefficiency of existing self-consistency based reasoning methods that rely on pre-sampling or additional model calls for difficulty estimation. We demonstrated that feed-forward network (FFN) activation patterns in large language models correlate with problem difficulty, and this information can be extracted without generating additional tokens. Based on this observation, we proposed Activation-Informed Difficulty-Aware Self-Consistency (ACTSC), which estimates problem difficulty using activations from a single forward pass and dynamically applies self-consistency sampling only when necessary. By using a lightweight activation-based probe, ACTSC enables adaptive inference without requiring a preparation stage. Experimental results across diverse reasoning benchmarks and multiple model scales demonstrate that ACTSC substantially reduces inference cost compared to existing self-consistency variants while maintaining comparable or even superior accuracy in some cases. These findings show that internal activations can serve as an effective signal for difficulty-aware test-time scaling and suggest a new direction for efficient inference control in large language models.

\bibliographystyle{named}
\bibliography{ijcai26}

\clearpage

\end{document}